\documentclass{article}

\usepackage{microtype}
\usepackage{graphicx}
\usepackage{subfigure}
\usepackage{booktabs} 

\usepackage{hyperref}
\usepackage{color}
\usepackage[table]{xcolor}
\usepackage[normalem]{ulem}

\usepackage[accepted]{icml2024}

\usepackage{xurl} 
\usepackage{hyperref}
\usepackage{graphicx}
\usepackage{array}
\usepackage{changepage}
\usepackage{multicol}
\usepackage{ragged2e}
\usepackage{pdflscape}
\usepackage{tabularray}
\usepackage[longtable]{multirow}
\usepackage{longtable}

\usepackage{caption}
\usepackage[pass]{geometry}
\usepackage{cleveref}
\usepackage{scalerel,xparse}

\definecolor{LFive}{rgb}{0.278, 0.365, 0.125}
\definecolor{LFour}{rgb}{0.584, 0.494, 0.039}
\definecolor{LThree}{rgb}{0.773, 0.361, 0.063}
\definecolor{LTwo}{rgb}{0.757, 0.51, 0.047}
\definecolor{LOne}{rgb}{0.753, 0.0, 0.0}
\definecolor{LNA}{rgb}{0.263, 0.263, 0.263}
\definecolor{LUnknown}{rgb}{0.6, 0.6, 0.6}
\definecolor{CompanyColor}{rgb}{0.851, 0.824, 0.914}
\definecolor{NonProfitColor}{rgb}{0.918, 0.82, 0.863}
\definecolor{GovernmentColor}{rgb}{0.812, 0.886, 0.953}

\newcommand{\CFiveTable}{\color{white}\cellcolor{LFive}C5}
\newcommand{\CFourTable}{\color{white}\cellcolor{LFour}C4}
\newcommand{\CThreeTable}{\color{white}\cellcolor{LThree}C3}

\newcommand{\COneTable}{\color{white}\cellcolor{LOne}C1}

\newcommand{\DFiveTable}{\color{white}\cellcolor{LFive}D5}
\newcommand{\DFourTable}{\color{white}\cellcolor{LFour}D4}
\newcommand{\DThreeTable}{\color{white}\cellcolor{LThree}D3}
\newcommand{\DTwoTable}{\color{white}\cellcolor{LTwo}D2}
\newcommand{\DOneTable}{\color{white}\cellcolor{LOne}D1}

\newcommand{\NATable}{\color{white}\cellcolor{LNA}N/A}
\newcommand{\UnkTable}{\color{white}\cellcolor{LUnknown}?}

\newcommand{\midrulenospacing}{\specialrule{.4pt}{2pt}{0pt}}
\newcommand{\bottomrulenospacing}{\specialrule{.8pt}{0pt}{2pt}}

\newcommand{\DevCompany}[1]{\cellcolor{CompanyColor}#1}
\newcommand{\DevNonProfit}[1]{\cellcolor{NonProfitColor}#1}
\newcommand{\DevGovernment}[1]{\cellcolor{GovernmentColor}#1}

\NewDocumentCommand\emojiplus{}{
    \scalerel*{
        \includegraphics{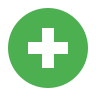}
    }{X}
}

\NewDocumentCommand\emojiminus{}{
    \scalerel*{
        \includegraphics{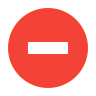}
    }{X}
}

\newcommand{\positiveImpact}[1]{\emojiplus #1}
\newcommand{\negativeImpact}[1]{\emojiminus #1}

\icmltitlerunning{Position: Near to Mid-term Risks and Opportunities of Open-Source Generative AI}

\begin{document}

\twocolumn[
\icmltitle{Near to Mid-term Risks and Opportunities of Open-Source Generative AI}



\icmlsetsymbol{equal}{*}

\begin{icmlauthorlist}
\icmlauthor{Francisco Eiras}{oxford}
\icmlauthor{Aleksandar Petrov}{oxford}
\icmlauthor{Bertie Vidgen}{mlcommons}
\icmlauthor{Christian Schroeder de Witt}{oxford}
\icmlauthor{Fabio Pizzati}{oxford}
\icmlauthor{Katherine Elkins}{kenyon}
\icmlauthor{Supratik Mukhopadhyay}{lsu}
\icmlauthor{Adel Bibi}{oxford}
\icmlauthor{Botos Csaba}{oxford}
\icmlauthor{Fabro Steibel}{its}
\icmlauthor{Fazl Barez}{oxford}
\icmlauthor{Genevieve Smith}{berkeley}
\icmlauthor{Gianluca Guadagni}{uov}
\icmlauthor{Jon Chun}{kenyon}
\icmlauthor{Jordi Cabot}{list,uol}
\icmlauthor{Joseph Marvin Imperial}{bath,nup}
\icmlauthor{Juan A. Nolazco-Flores}{itesm}
\icmlauthor{Lori Landay}{berklee}
\icmlauthor{Matthew Jackson}{oxford}
\icmlauthor{Paul Röttger}{bocconi}
\icmlauthor{Philip H.S. Torr}{oxford}
\icmlauthor{Trevor Darrell}{berkeley}
\icmlauthor{Yong Suk Lee}{notredame}
\icmlauthor{Jakob Foerster}{oxford}
\end{icmlauthorlist}

\icmlaffiliation{oxford}{University of Oxford}
\icmlaffiliation{mlcommons}{MLCommons}
\icmlaffiliation{lsu}{Center for Computation \& Technology, Louisiana State University}
\icmlaffiliation{bath}{University of Bath}
\icmlaffiliation{nup}{National University Philippines}
\icmlaffiliation{bocconi}{Bocconi University}
\icmlaffiliation{berkeley}{University of California, Berkeley}
\icmlaffiliation{list}{Luxembourg Institute of Science and Technology}
\icmlaffiliation{uol}{University of Luxembourg}
\icmlaffiliation{notredame}{University of Notre Dame}
\icmlaffiliation{kenyon}{Kenyon College}
\icmlaffiliation{berklee}{Berklee College of Music}
\icmlaffiliation{its}{Institute for Technology \& Society (ITS), Rio}
\icmlaffiliation{uov}{University of Virginia}
\icmlaffiliation{itesm}{ITESM}

\icmlcorrespondingauthor{FE}{eiras@robots.ox.ac.uk}

\icmlkeywords{Machine Learning, ICML}

\vskip 0.3in
]

\printAffiliationsAndNotice{}

\begin{abstract}
In the next few years, applications of Generative AI are expected to revolutionize a number of different areas, ranging from science \& medicine to education. 
The potential for these seismic changes has triggered a lively debate about potential risks and resulted in calls for tighter regulation, in particular from some of the major tech companies who are leading in AI development. 
This regulation is likely to put at risk the budding field of open-source Generative AI. We argue for the responsible open sourcing of generative AI models in the near and medium term.
To set the stage, we first introduce an AI openness taxonomy system and apply it to 40 current large language models. We then outline differential benefits and risks of open versus closed source AI and present potential risk mitigation, ranging from best practices to calls for technical and scientific contributions. 
We hope that this report will add a much needed missing voice to the current public discourse on near to mid-term AI safety and other societal impact.

\end{abstract}

\section{Introduction}
\vspace{-0.1cm}
\label{sec:introduction}

Generative AI (Gen AI), defined as \textit{``artificial intelligence that can generate novel content"} by conditioning its response on an input \citep{gozalo2023chatgpt} (e.g., large language or foundation models), is anticipated to profoundly impact a diverse array of domains including science \citep{ai4science2023impact}, the economy \citep{brynjolfsson2023generative}, education \citep{alahdab2023potential}, the environment~\citep{rillig2023risks}, among many others. As a result, there has been significant socio-technical work undertaken to evaluate the broader risks and opportunities associated with these models, in a step towards a more nuanced and comprehensive understanding of their impacts \citep{bommasani2021opportunities}, including recent regulatory developments (see Appendix ~\ref{app:ai_reg}).

\begin{figure}[t]
    \centering
    \includegraphics[width=0.9\linewidth]{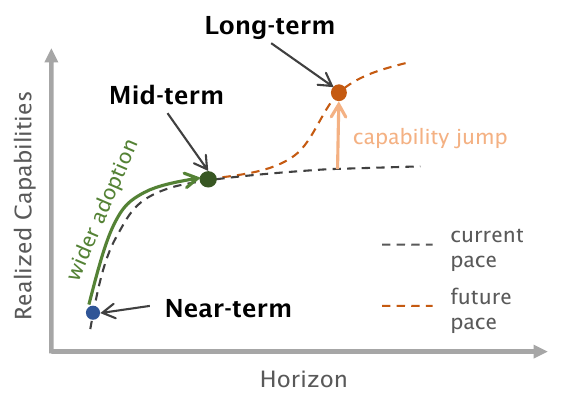}
    \vspace{-0.8em}
    \caption{\textbf{Three Development Stages for Generative AI Models}: \textit{near-term} is defined by early use and exploration of the technology in much of its current stage; \textit{mid-term} is a result of the widespread adoption of the technology and further scaling at current pace; \textit{long-term} is the result of technological advances that enable greater AI capabilities.}
    \label{fig:development-stages}
    \vspace{-0.60cm}
\end{figure}

Parallel to these efforts is a debate on the \textit{openness of Gen AI} models. The digital economy heavily relies on open-source software, exemplified by over 60\% of global websites using open-source servers like Apache and Nginx \citep{lifshitz2021digital}. This prevalence is underscored by a 2021 European Union report, which concluded that ``overall, the [economic] benefits of open source greatly outweigh the costs associated with it'' \citep{blind2021impact}. Some developers of Gen AI models have chosen to openly release trained models (and sometimes data and code too), by leaning on this narrative and claiming that by doing so \textit{``[these models] can benefit everyone''} and that \textit{``it's safer [to release them]''} \citep{meta2023meta}. However, while there has been a flurry of reports and surveys on the impacts of general open-source software in areas such as innovation or research within the last few decades \citep{paulson2004empirical,schryen2009open,von2007open}, the discourse surrounding the openness of Gen AI models presents unique complexities due to the distinctive characteristics of this technology, including e.g., potential dual use and run-away technological progress.

This paper argues that the success of open source in traditional software could be replicated in Gen AI with well-defined and followed principles for responsible development and deployment. To this end, we begin by defining different stages of Gen AI development/deployment, followed by an empirical analysis of the openness of existing models through a taxonomy. With this framework, we then focus on evaluating the risks and opportunities presented by open and closed source Gen AI in the near to mid-term. Finally, we make a case for \textbf{the responsible open sourcing of generative AI models developed in the near to mid-term stages}, presenting recommendations to developers on how to achieve this safely and efficiently.

\begin{figure}[t]
    \centering
    \includegraphics[width=\linewidth]{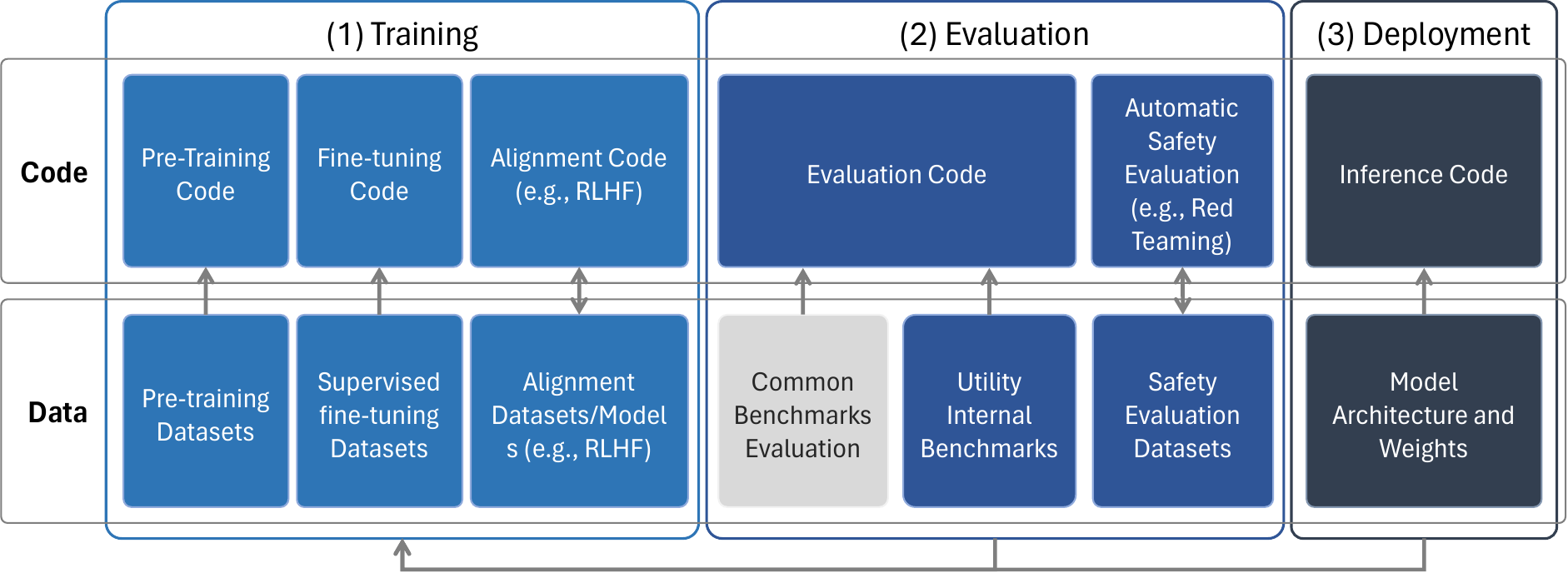}
    \caption{\textbf{Model Pipeline}: stages showing (1) training, (2) evaluation, and (3) deployment analyzed in the report. The component Common Benchmarks Evaluation (light gray) is included for completeness yet will not be analyzed in detail as these are standard and commonly available.}
    \label{fig:openness-pipelines}
    \vspace{-0.60cm}
\end{figure}

\section{Preliminaries}
\vspace{-0.1cm}
\label{sec:preliminaries}

To frame our analysis of the impacts of open sourcing generative AI models, we start by defining three-stages of AI development and outline the current pipelines involved in training, evaluating and deploying Large Language Models (LLMs). We focus on LLMs in these definitions and in \S \ref{sec:taxonomy} as this is the modality with the most prolific model development and open-sourcing at the moment, but note that it would be easy to extend our analysis to other modalities.

\textbf{Stages of Development of Gen AI Models} Our three-part framework (Figure \ref{fig:development-stages}) to describe the evolution of generative AI focuses on adoption rates and technological advancements instead of time elapsed (similar to \citeauthor{anthropic2023}, \citeyear{anthropic2023}). The \textbf{near-term} stage is defined by the early use and exploration of existing technology, such as deep learning with transformer and diffusion model architectures, utilizing large datasets. This phase is characterized by experimentation, with increasing levels of development, investment and adoption. The \textbf{mid-term} is defined by the widespread adoption and scaling of existing technology, and the exploitation of its benefits. We conceptualize this as moving along a predictable `capability curve', whereby more resources and usage will lead to greater benefits (and risks), but technological capabilities have not radically improved. Increasing use of multimodal models, agentic systems, and retrieval augmented generation are expected at this stage. The \textbf{long-term} is defined by a technological advance that will create dramatically greater AI capabilities, and therefore more risks and opportunities. This could manifest as a novel AI paradigm, a departure from traditional deep learning architectures, more efficient data utilization, among others, leading to more powerful AI models. In this paper, we focus primarily on analyzing the risks and opportunities of open-source Gen AI in the near to mid-term stages.

\textbf{Training, Evaluating, and Deploying LLMs}
The components typically involved in the (1) training, (2) evaluation, and (3) deployment of models are shown in Figure \ref{fig:openness-pipelines}, and they can be divided into two categories: \textit{Code} and \textit{Data}. 
We briefly describe each of the stages below, and provide a more in-depth component description in Appendix \ref{app:pipeline_details}.

Model training processes can be grouped into three distinct stages: \textit{pre-training}, where a model is exposed to large-scale datasets composed of trillions of tokens of data, with the goal of developing fundamental skills and broad knowledge; \textit{supervised fine-tuning} (SFT), which corrects for data quality issues in pre-training datasets using a smaller amount of high-quality data; and \textit{alignment}, focusing on creating application-specific versions of the model by considering human preferences. Once trained, models are usually evaluated on openly available evaluation datasets (e.g., MMLU by \citeauthor{hendrycks2020measuring}, \citeyear{hendrycks2020measuring})
as well as curated benchmarks (e.g., HELM by \citeauthor{liang2022holistic}, \citeyear{liang2022holistic}). Some models are also evaluated on utility-oriented proprietary datasets held internally by developers, potentially by holding out some of the SFT/alignment data from the training process \citep{touvron2023llama}. On top of utility-based benchmarking, developers sometimes create safety evaluation mechanisms to proactively stress-test the outputs of the model (e.g., red teaming via adversarial prompts). Finally, at the deployment stage, content can be generated by running the inference code with the associated model weights.

\begin{figure}[t]
    \centering
    \includegraphics[width=\linewidth]{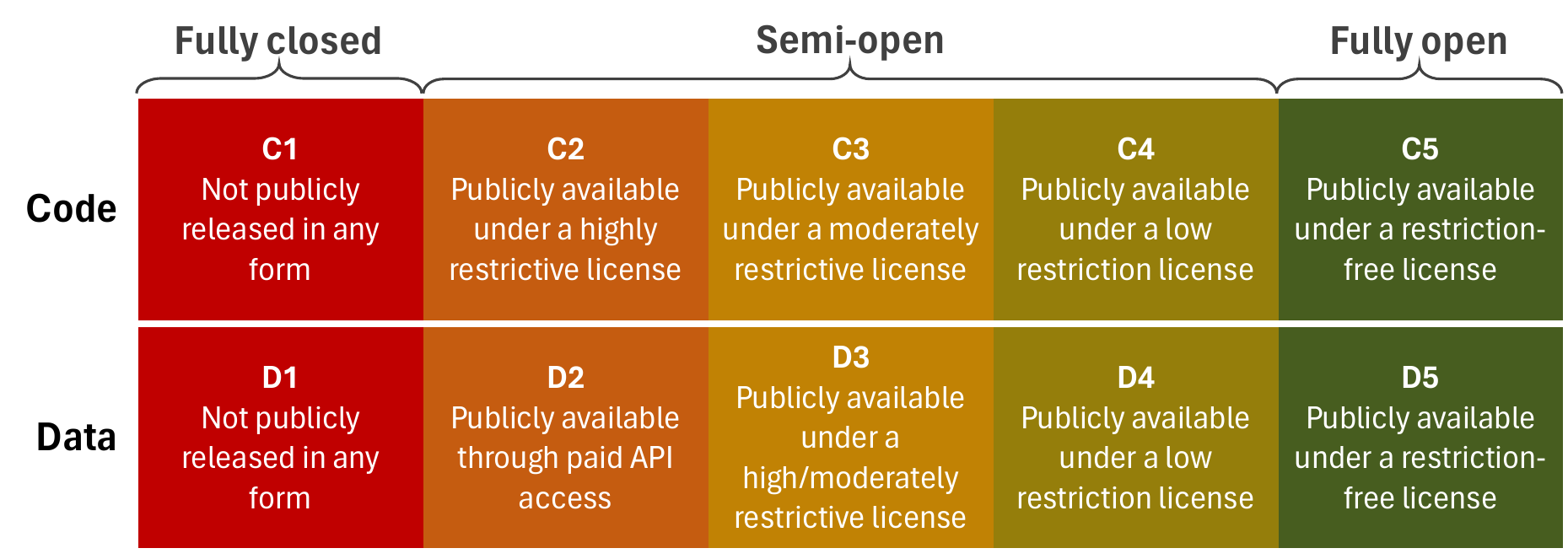}
    \vspace{-1.7em}
    \caption{\textbf{Openness Scale}: categorization of the levels of openness of the code and data of each model component. See Table \ref{tab:licenses_full} (Appendix \ref{app:taxonomy_tables}) for the restrictions of each license.}
    \label{fig:openness-levels}
    \vspace{-0.60cm}
\end{figure}

\section{Openness Taxonomy of LLMs}
\vspace{-0.1cm}

Model developers decide whether to make each component of the training, evaluation and deployment pipeline (Figure \ref{fig:openness-pipelines}) \textit{private} or \textit{public}, with varying levels of restrictions for the latter. 
For instance, the developers of LLaMA-2 have publicly released the model architecture and weights, yet they have not shared the code or reward model for Reinforcement Learning from Human Feedback (RLHF) used in the Alignment components \citep{touvron2023llama}. 
To properly evaluate the openness of each component, we introduce a classification scale for Gen AI models in \S \ref{sec:openness_classification}, which we then apply to 40 high impact LLMs in \S \ref{sec:taxonomy}. This will help contextualizing the risks and opportunities discussed in \S \ref{sec:near-to-mid-risks}, and the responsible open sourcing argument we make in \S \ref{sec:recommendations}.
An up-to-date version of the taxonomy of LLMs is also available on \href{https://open-source-llms.github.io}{this link}.

\subsection{Classifying Openness for Gen AI Code and Data}
\vspace{-0.1cm}
\label{sec:openness_classification}

We introduce a framework for categorizing the openness of each component of Gen AI pipelines (e.g., Figure \ref{fig:openness-pipelines}). At the highest level, a \textbf{fully closed} component is not publicly accessible in any form \citep{rae2022scaling}. In contrast, a \textbf{semi-open} component is publicly accessible but with certain limitations on access or use, or it is available in a restricted manner, such as through an Application Programming Interface (API) \citep{achiam2023gpt}. Finally, a \textbf{fully open} component is available to the public without any restrictions on its use \citep{xu2022systematic}.
Further, the semi-open category comprises three subcategories, delineating varied openness levels (see Figure \ref{fig:openness-levels}). Distinctions are made between Code (C1-C5) and Data (D1-D5) components, where C5/D5 represents unrestricted availability and C1/D1 denotes complete unavailability. 
For semi-open components, their classification relies on the license of the publicly available code/data. 

\begin{figure}[t]
    \centering
    \includegraphics[width=0.98\linewidth]{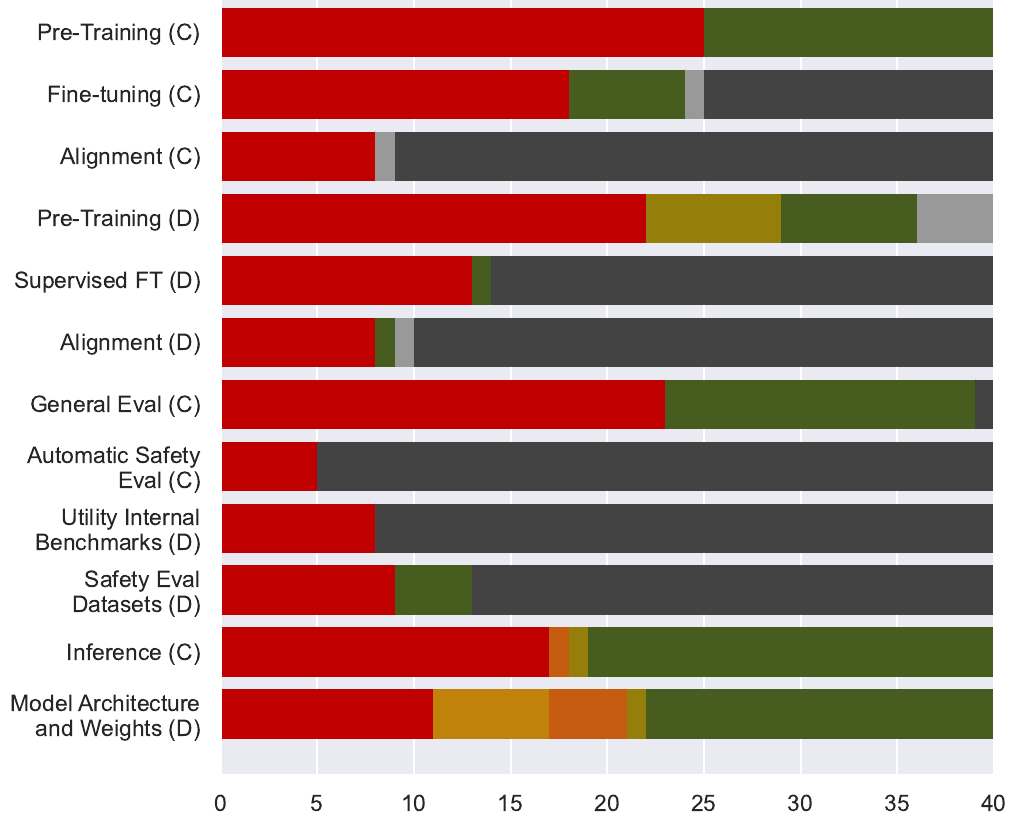}
    \vspace{-1.0em}
    \caption{\textbf{Distribution of Openness Levels by Pipeline Component}: openness level distribution for each of the pipeline components of the 40 LLMs studied. Color legend: \colorbox{LOne}{\color{white} C1/D1}, \colorbox{LTwo}{\color{white} C2/D2}, \colorbox{LThree}{\color{white} C3/D3}, \colorbox{LFour}{\color{white} C4/D4}, \colorbox{LFive}{\color{white} C5/D5}, \colorbox{LUnknown}{\color{white} ?} (unknown or not publicly available), \colorbox{LNA}{\color{white} N/A} (not applicable). For conciseness, we use "FT" as a stand in for "Fine-Tuning".}
    \label{fig:classification_distribution}
    \vspace{-0.60cm}
\end{figure}

To evaluate the licenses we introduce a point-based system where each license gets 1 point (for a total maximum of 5) for allowing each of the following: \textit{can use a component for research purposes} (\textbf{Research}), \textit{can use a component for any commercial purposes} (\textbf{Commercial Purposes}), \textit{can modify a component as desired (with notice)} (\textbf{Modify as Desired}), \textit{can copyright derivative} (\textbf{Copyright Derivative Work}), \textit{publicly shared derivative work can use another license} (\textbf{Other license derivative work}).
The total number of points is indicative of a license's restrictiveness. A \textbf{Highly restrictive} license scores 0-1 points, aligning with openness levels of code C2 and data D3, imposing significant limitations. A \textbf{Moderately restrictive} license, scoring 2-3 points (code C3 and data D3), allows more flexibility but with some limitations. Licenses scoring 4 points are \textbf{Slightly restrictive} (code C4 and data D4), offering broader usage rights with minimal restrictions. Finally, a \textbf{Restriction free} license scores 5 points, indicating the highest level of openness (code C5 and data D5), permitting all forms of use, modification, and distribution without constraints.

In Table \ref{tab:licenses_full} (Appendix \ref{app:taxonomy_tables}) we provide a full table with the openness licenses and levels of all models studied in \S \ref{sec:taxonomy}. 

\subsection{Openness Taxonomy of Current LLMs}
\vspace{-0.1cm}
\label{sec:taxonomy}

We analyzed the pipeline components of 40 high-impact LLMs released from 2019 to 2023, chosen by optimizing three key impact metrics: \textit{ChatBot Arena Elo Rating}, a crowdsourced benchmark score comparing models\footnote{Introduced in 05/2023; older models may be underrepresented.}; \textit{Google Scholar Citations}, indicating each model's academic impact; and \textit{HuggingFace Downloads Last Month}, reflecting the usage of models openly available on HuggingFace. While we included models that scored high on any of these metrics, we also decided to include other released models for the sake of diversity. Due to space constraints, the full model list is in Table \ref{tab:model_list_full} (Appendix \ref{app:taxonomy_tables}). 

A full table with the taxonomy of each of the model components is presented in Table \ref{tab:classification_full} (Appendix \ref{app:taxonomy_tables}). In Figure \ref{fig:classification_distribution}, we show the distribution of openness levels for each of the pipeline components analyzed. 
Figure \ref{fig:classification_distribution} clearly shows a balance between open and closed source deployed components (inference code and weights); however, \textit{\textbf{a notable skew exists towards closed source in training data (such as fine-tuning and alignment) and, importantly, in safety evaluation code and data}}. 
To fully leverage open source benefits and mitigate risks discussed in the next sections, a significant shift toward responsible development and deployment of open-source generative AI is necessary.

\section{Near to Mid-term Risks and Opportunities of Open Source Gen AI Models}
\vspace{-0.1cm}
\label{sec:near-to-mid-risks}

We describe the risks  and opportunities provided by open-source models in the near and mid-term (as defined in \S \ref{sec:preliminaries}). 
Our focus is how open source catalyses, minimizes or creates risks and benefits compared to closed source -- rather than Gen AI in general. 
Unless stated explicitly, we refer to all artifacts and components of AI when using the term ``open source''. 

\textbf{The Challenges of Assessing Risks and Benefits} 
Gen AI systems can be evaluated through a variety of methods and frameworks, such as benchmarks like HELM and Big-Bench for task evaluation, Chatbot Arena for crowd-sourced model comparisons, and red teaming for exploratory evaluation \cite{guo2023evaluating, liang2023holistic, srivastava2023imitation}. However, these approaches face limitations like limited ecological validity and data contamination \cite{li2023static, sainz2023nlp, zhou2023dont}, and provide only a partial view of how models will perform in real-world settings. 
In response, some experts suggest socio-technical evaluations that are focused on real-world applications \cite{weidinger2023,solaiman2023evaluating}. This is supported by calls for comprehensive pre-release audits of models, datasets, and research artifacts \cite{derczynski2023assessing, M_kander_2023, 10.1145/3600211.3604712}. 
However, even 
holistic approaches to evaluation face substantial challenges, such as the rapid and unpredictable evolution of AI capabilities, the difficulty of standardizing measurements due to the fast pace of change, and the research community's limited insight into AI's industrial applications. 
This invariably leads to partial and incomplete evidence.
As such, while we use diverse evidence to examine and support our arguments, it is important to recognize the challenges in reaching definitive conclusions as a result of these limitations.

\subsection{Quality and Transparency}
\vspace{-0.1cm}
\label{sec:quality_and_transparency}

\textbf{\positiveImpact{Open Models are More Flexible and Customizable}}
Having access to open-source models, datasets, and assets significantly aids developers in creating models that are high-performing and specifically tailored to their use-case. 
Developers have access to far more training approaches, models and datasets. This gives them a powerful starting point when creating a model for a specific application. It also particularly helps cater to less well-resourced languages, domains, and downstream tasks \cite{bommasani2}, as well as enabling personalized models that cater to distinct groups and individuals \cite{kirk2023personalisation}.
This has created widespread positive sentiment towards open source, which can be seen in venture capital firm's significant investment in open-sourcing efforts \citep{bornstein2023supporting, horowitz}, and the growing adoption of open-source models by companies \citep{marshall2024how}.

\textbf{\positiveImpact{Open Source Improves Public Trust Through More Transparency}}
Nearly three out of five people (61\%) are either ambivalent about or unwilling to trust AI, with \citet{gillespie2023trust} reporting that cybersecurity risks, harmful use, and job loss are the ``potential risks'' that people are most concerned about. 
Closed source models pose challenges for evaluating,
benchmarking, and testing them which impede accessibility, replicability, reliability, and
trustworthiness \citep{lamalfa2023language}.
Transparency is a powerful way of improving trust, and addressing this critical problem. 
Transparency includes providing clear and explicit documentation, such as provenance artefacts like model cards, datasheets, and risk cards \cite{gebru2021datasheets, derczynski2023assessing, longpre2023data}. They can be used to assess and review datasets and models, and are widely-used in the open source community. 
Open source is itself the best way of creating transparency. It enables widespread community oversight as models and datasets can be interrogated, scrutinised, and evaluated by anyone, without needing to seek approval from a central decision-maker. This empowers developers, researchers and other actors to engage with AI and contribute to discussions, encouraging a culture of contribution and accountability \cite{sanchez2021civil}. 
At the same time, the highly technical nature of AI research creates substantial barriers to typical citizens. As such, more transparency may not alone drive greater trust -- research outputs also need to be \textit{accessible} and \textit{understandable} by non-experts \cite{Mittelstadt_2019}.

\subsection{Research and Academic Impact}
\vspace{-0.1cm}
\label{sec:research}

\textbf{\positiveImpact{Open Source Advances Research}}
Compared to the machine learning landscape a decade ago, the availability and continuous growth of open source in recent years has enabled the community to do more diverse and innovative research. This includes researchers exploring the inner workings of models through jailbreaking and quality checking for unsafe, harmful, and biased content (see \S \ref{sec:safety}) as well as probing for misuse of copyrighted data, which can potentially lead to class-action lawsuits (see \S \ref{sec:societal_impact}). 
Likewise, the availability of code, data, and proper documentation of open models have allowed researchers to develop novel breakthroughs (e.g., DPO \cite{rafailov2023direct} as a more cost-efficient substitute for RLHF \cite{ouyang2022training} for capturing human preference), which have been proven to boost open models to gain comparable performances against their closed model counterparts.
Closed models, on the other hand, only grant limited access through API calls and restrict access to essential model generation outputs such as logits and token probabilities. Such limitations restrict researchers from forming deeper methodological insights and limit reproducibility of their research \cite{rogers-closed}.

\subsection{Innovation, Industry and Economic Impact}
\label{sec:economic_diversity}
\vspace{-0.1cm}

\textbf{\positiveImpact{Open Source Empowers Developers and Fosters Innovation}}
Closed source models accessed via an API make product developers reliant on an external provider for essential components of their product or system. This reliance can limit control and maintainability, especially as models can be updated or removed without warning by their owners. 
Further, with a closed model developers may not own their data or have full control over their data pipeline, which can make it more difficult to innovate on design, steer model performance, change aspects of their system, or understand their own workflows. 
In contrast, open models offer significant advantages. Developers can modify the model according to their needs, have complete understanding and transparency of the model, and control the data pipeline, which greatly enhances privacy and auditability \cite{culotta2023use}. One important consideration is whether models are released with permissive licenses that suit commercial usecases (see commercial use in \S \ref{sec:preliminaries}). This is increasingly common with more recent releases.
Open-source models could be particularly beneficial in the emerging field of generative AI-powered agents \cite{chan2024visibility}, where outputs involve performing digital or physical actions (for early examples see Adept's blog post \cite{adepts_post}, and Amazon's press release \cite{amazon_press_release}). In this context, product developers are likely to value having more control over models, being able to deploy them on-device, and integrate them in larger, more complex systems. 

\textbf{\positiveImpact{Open Source Can be More Affordable}}
AI models can enhance individual productivity by automating repetitive and time-consuming tasks, and augmenting workers when completing more complex and high-value tasks. This can help narrow the productivity gap between workers, improving minimum performance standards \cite{dell2023navigating}. 
In principle, open-source AI models increase these benefits as they are available for free. However, substantial operational costs are still involved, such as the staff required to run the models, the time of leadership to organise and oversee their use, and the compute costs for inference \citep{palazzolo2023meta}. 
Some enterprises might also apply additional protections for security and data to ensure compliance when using open-source models, adding further costs. 
Whether open source is cheaper overall than closed source depends on the maturity and capabilities of the organisation. Generally, larger corporations can bear the overheads involved in open source and overall make substantial savings. 

\textbf{\positiveImpact{Open Source Can be Easier to Access}}
Open-source models are increasingly easy to use and access, with a range of vendors providing SDKs, APIs and downloadable files, such as \href{https://replicate.com/}{Replicate}, \href{https://www.together.ai/}{Together}, and \href{https://huggingface.co/}{HuggingFace}.
Further, they typically require few approvals to start using models, in comparison with more onerous signup processes from closed source providers.
One important area where open source lags behind closed source is in providing user interfaces aimed at non-technical audiences. While ChatGPT is easy to interact with and well-known amongst the general public, few open-source models have widely-used UIs.

\textbf{\positiveImpact{Open Source Could Achieve Comparable Performance}}
Today, the preference for closed source models stems from their user-friendly packaging, cost-effectiveness (with lower-income individuals predominantly opting for free versions, see \citeauthor{mollick}, \citeyear{mollick}), and potentially superior performance across various tasks (\href{https://huggingface.co/spaces/HuggingFaceH4/open_llm_leaderboard}{Open LLM Leaderboard}). However, these dynamics are likely to shift in the near to mid-term. Firstly, with the growth of open source development, the performance gap between open and closed source models is expected to narrow significantly \cite{uk_gov_safety_2025}. Further, open source might be better in specific applications and contexts (see \S \ref{sec:economic_diversity}), driving adoption. 

\textbf{\positiveImpact{Open Models Could Help Tackle Global Economic Inequalities}}
Knowledge workers in low-income nations, including workers in sectors like call centers and software development, face serious risk of job losses as AI models automate and semi-automate their work. 
Further, if AI models fail to adapt to local contexts or remain financially inaccessible, the expected economic benefits and new job opportunities may not arise, worsening economic inequalities \cite{imf_blog}.
This is a concern as closed source models are often (1) unaffordable for companies in low-income countries and (2) badly-suited to their needs (see \S \ref{sec:societal_impact}). Local needs are often not met because they lack adequate language support, culturally relevant content, and effective safety measures. This results in higher costs and lower performance, compounding the global inequalities that could be caused by generative AI \cite{petrov2023language, ahia2023all}. 
In contrast, open models could significantly change this dynamic. With requisite skill building and support for different communities, open models would enable communities to tailor models to their specific contexts and needs, promoting local innovation, safety, security, and reduced bias. This shift could help bridge the growing global inequality gap, paving the way for a more equitable and inclusive future in generative AI.

\subsection{Safety}
\vspace{-0.1cm}
\label{sec:safety}
Generative AI models can create safety risks by increasing the severity and prevalence of harm experienced by individuals and society at large. This can take many forms, including physical, psychological, economic, representational and allocational harms \cite{shelby2023sociotechnical, weidinger2023}.
The primary risks from current and near-term generative AI capabilities comprise two distinct pathways. 
The first is \textit{malevolent use by bad actors}: individuals or organizations might exploit AI to create damaging content or enable harmful interactions, such as personalized scams, targeted harassment, sexually explicit and suggestive content, and disinformation on a large scale \citep{vidgen2023simplesafetytests,ferrara2023genai}.
The second is \textit{misguidance of vulnerable groups}: inaccurate or harmful advice from AI could lead vulnerable individuals, including those with mental health issues, to engage in self-harm \citep{mei2022mitigating,mei2023assert,rottger2023xstest}, radicalise towards supporting extremist groups, or believe in factually inaccurate claims about elections, health, and the environment \citep{zhou2023synthetic}. 
In the long-term, AI might develop capabilities that present novel existential threats, creating ``catastrophic'' consequences for society such as chemical warfare and environmental disaster \cite{hendrycks2023overview, shevlane2023model, matteucci2023ai}. However, these risks are not a substantial concern for existing models given their limited capabilities. 
Thus, in the near to mid-term, AI safety primarily means preventing models from generating toxic content, giving dangerous advice, and following malicious instructions.

\textbf{\positiveImpact{Open Source Enables Technological Innovation for Safety}}
Open source has significantly advanced safety research in the entire model development pipeline. Large open datasets for pre-training, like the Pile \citep{gao2020pile} (released for GPT-Neo, studied in the taxonomy \S \ref{sec:taxonomy}), Laion \citep{schuhmann2022laion}, and RedPajama \citep{together2023redpajama}, can be analysed for whether they contain toxic content \cite{prabhu2020large}. 
Similarly, open research has shown model fine-tuning to be highly efficient in both improving model safety and removing model safeguards \citep[e.g.][]{bianchi2023safety,qi2023fine}. Unlike closed APIs, open model analyses permit in-depth exploration of internal mechanisms and behaviors \citep[e.g.][]{jain2023mechanistically, casper2024blackbox}. This transparency enables reproducible and comprehensive evaluations, strengthening our understanding of generative AI safety for models with near and mid-term capabilities. 
Open source has also driven innovation in developing safeguards and controls for models, such as Meta's LlamaGuard \cite{inan2023llama} and HuggingFace's \href{https://huggingface.co/spaces/AI-Secure/llm-trustworthy-leaderboard}{Safety Evaluation Leaderboard}.

\textbf{\negativeImpact{Open Models Can Also be Made to Generate Unsafe Content}}
The flexibility of open-source models, as discussed in \S \ref{sec:quality_and_transparency}, has its drawbacks. Despite their initial alignment, these models can be fine-tuned to produce unsafe content, as exemplified by \href{https://huggingface.co/ykilcher/gpt-4chan}{GPT4Chan} and various \href{https://huggingface.co/models?search=uncensored}{``uncensored models''} on the HuggingFace hub, designed to execute any instruction, irrespective of its safety implications. It is important to recognize, however, that closed models are not impervious to similar risks. Jailbreaks can induce unsafe behaviors in closed models as well \citep{zou2023universal}, and recent studies have demonstrated that closed models can easily be fine-tuned to become just as unsafe as open ones \citep{qi2023fine}. Nonetheless, ongoing advancements in generative AI safety technology \citep{dai2023exploring}, particularly through open models, hold the potential for mitigating these risks in the near to mid-term horizon.

\textbf{\negativeImpact{Open Models Cannot be Rolled Back or Updated}}
Once a model is made public, anyone can download it and use it indefinitely. In principle, benign users' access (e.g., researchers or rule-abiding corporations) can be regulated through license modifications. However, not all benign users will be aware of license changes and malicious actors will choose to not follow them. This creates a safety risk as any problems that have been identified post-deployment cannot be addressed.
In comparison, closed model developers can cut off access to unsafe models if they are gatekept through an API. 
To reduce these risks, open source developers and the communities that host models (e.g., HuggingFace) must adhere to responsible release and access policies (e.g. \citealt{solaiman2023gradient, solaiman2023evaluating, anthropic2023}).

\subsection{Societal and Environmental Impact}
\vspace{-0.1cm}
\label{sec:societal_impact}

\textbf{\positiveImpact{Open Source Models Can Reduce Energy Use}}
AI model training incurs significant environmental costs from the energy consumption of compute resources. \citep{strubell2019energy, wu2022sustainable}.
These impacts, measurable in CO$_2$ emissions, span the entire AI development process, including training and inference \citep{verdecchia2023systematic, kumar2023how}.
While accurately quantifying emissions for cloud providers is challenging due to variables like hardware utilization, team practices, geography, and time of day, industry-wide energy consumption can be reduced by sharing of resources that are energy-intensive to create, such as model weights \citep{saenko2023computer}. 
In addition, open-sourcing can lead to transparent profiling of code to identify energy bottlenecks. This can then be addressed by the community, creating more energy-efficient training methods. For instance, some researchers have put forward small model development paradigms \cite{schwartz2019green}.

\textbf{\positiveImpact{Open Models Can Help With Copyright Disputes}}
One of the major legal issues surrounding generative AI is the use of copyrighted data for training without explicit permission \citep{llmlitigation,metz2024openai}. This has mostly been identified because models regurgitate memorized data when prompted in specific ways \cite{karamolegkou-etal-2023-copyright,carlini2022quantifying}. The lack of transparency about what data are used in model training for both open and closed source (highlighted in \S \ref{sec:taxonomy}) can lead to confusion, uncertainty, and misattribution. 
Open models that release, or describe, their training data can help address these issues of data privacy, memorization and the ``fair use'' of copyrighted materials.
Crowd-sourced data curation also offers a way of minimizing use of proprietary datasets in the future, reducing the risk of copyright disputes \cite{hartmann2023sok}. 

\textbf{\positiveImpact{Open Models Can Serve the Needs and Preferences of Diverse Communities}}
To address global needs effectively, it is crucial that models do not only reflect the values of people who are liberal, culturally Western, and English speaking \citep{aroyo2023dices, lahoti2023improving}. 
However, models are largely trained on data from the Internet, which is often biased to such people \citep{joshi2020state}. 
There is a pressing need to make pre-training datasets more diverse, inclusive and representative.
In the short-term, models can be \textit{steered} to meet the needs of different contexts, languages, and communities. Open source is a powerful way of achieving this as it enables under-resourced actors to build on top of each other's contributions.
For instance, platforms like HuggingFace host a vast array of models, with many designed for specific cultural, geographic, or linguistic needs, e.g., Latxa \cite{bandarkar2023belebele} and LeoLM \cite{hessianai-7b-chat}, covering diverse domains \citep[e.g.][]{li2023chatdoctor}.

\textbf{\positiveImpact{Open Source Helps Democratize AI Development}}
Open source empowers developers to utilize resources from major organizations (e.g., companies, governments or research labs), facilitating the reuse of assets and leading to time, effort and money savings. This is crucial for AI development, which is characterized by high costs and complexity, from pre-training models that can cost millions \cite{knight2023openai} to the creation of expensive human-labeled datasets.
This creates a clear societal benefit by enabling non-elites to access and use AI, which can include creating economic opportunities (see \S \ref{sec:economic_diversity}). It is important to acknowledge that, at a higher level, open-source models still contain key decisions, datasets and approaches that influence what is built on top of them. In this sense, they are currently undemocratic. They are informed by the values and market priorities of their largely for-profit driven developers.

\section{Responsible Open Sourcing of Near to Mid-Term Generative AI}
\vspace{-0.1cm}
\label{sec:recommendations}

\subsection{Addressing Common Concerns on Open Sourcing Generative AI}
\vspace{-0.1cm}
Despite the many benefits of open source, concerns surrounding the increased potential for malicious use, and uncertainty about its societal impact, have prompted calls for keeping generative AI closed source \citep{seger2023open}. There are real risks associated with open-source models. However, we believe these are sometimes exaggerated, possibly motivated by the economic interests of market leaders. Most concerns about open sourcing near to mid-term AI models are also pertinent to closed source models.

\textbf{CLAIM \#1: Closed Models Have Inherently Stronger Safeguards than Open-Source Models}
Several studies demonstrate that closed models typically demonstrate fewer safety and security risks, compared to open source \cite{rottger2023xstest, chen2024chatgpts, sun2024trustllm}. 
However, closed models still demonstrate weaknesses, and are particularly vulnerable to jailbreaking techniques \citep{zou2023universal,chao2023jailbreaking}.
Closed model safeguards are easily bypassed through simple manipulations like fine-tuning via accessible services \citep{qi2023fine}, prompting the model to repeat a word \citep{nasr2023scalable}, applying a cypher \cite{yuan2023gpt4}, or instructing the model in another language \citep{deng2023multilingual,yong2023low}.
Completely preventing models from exhibiting undesirable behaviors might not even be possible \citep{wolf2023fundamental,petrov2024prompting}.
Therefore, it is not clear that closed models are definitively ``safer'' than open-source models. We also anticipate that gaps will narrow over time as open safeguarding methods continue to improve.

\textbf{CLAIM \#2: Access to Closed Models Can Always be Restricted }
Closed models are often considered more secure because access can be restricted or removed if problems are identified. However, closed models can be compromised via hacking, leaks \citep{llamaLeak}, reverse engineering \citep{appleHack} or duplication \citep{Oliynyk2023stealing}. 
This perspective also assumes that models are only offered through an API. But some closed models are delivered on premise/device, particularly for sensitive deployments (e.g., government applications). In such cases, access may not be retractable.
Finally, closed models can be leaked, e.g., Mistral's 70B parameter was leaked by one of their early customers \citep{franzen2024}. 
Given these factors, developers do not always have the ability to unilaterally revoke access.

\textbf{CLAIM \#3: Closed Source Developers Can be Regulated to be Safer }
Regulatory pressure is primarily aimed at large companies building closed source models (e.g., see \href{https://www.whitehouse.gov/briefing-room/statements-releases/2023/10/30/fact-sheet-president-biden-issues-executive-order-on-safe-secure-and-trustworthy-artificial-intelligence/}{White House Executive Order}). 
While it can create incentives for safe model development, regulation is not a panacea, and several closed source models have been released that are uncensored, poorly safeguarded \citep{elevenScam} or deliberately misaligned \citep{wormGPT,Grok1,Grok2}.
It is also not clear that regulating closed source models is an effective way of stopping malicious actors \citep{yakuza,NigerianScammers}, who are capable of creating and distributing their own closed source models via illicit sales channels \citep{AiCybercriminal}. 
Instead, it might create higher costs for legitimate users who are restricted in what models they can access \citep{wu2023llmdet}.

\textbf{CLAIM \#4: All Safety and Security Problems Must be Addressed By the Model Provider }
It is becoming increasingly clear that, because of the numerous potential applications of generative models, all safety risks cannot be simply identified (and stopped) by the model provider. 
First, most model risks depend on the context and actors, and their real-world resources. For instance, real-world constraints significantly hinder activities like acquiring chemicals, equipment, or weapons, thus limiting open source's potential for misuse in such endeavors. 
Second, models may not have a causal impact on actors if they either (a) have other means of inflicting harm -- such as searching on the web for malicious information -- or (b) pay little attention to the responses of the model.
Third, in practice, other stakeholders help protect people from risk through established safeguarding practices, such as Internet Service Providers, cloud services, social media, and law enforcement. 
Given these factors, safety and security issues cannot be seen as solely the responsibility of the model provider.

\subsection{Recommendations for Safe and Responsible Open Sourcing of Near to Mid-term Gen AI Models}
\label{sec:recs}
\vspace{-0.1cm}
To safely and responsibly open-source Gen AI models, we outline five important priorities for developers, starting with technical recommendations ahead of broader responsibility and socio-technical considerations.

\textbf{Enhance Data Transparency and Provenance }
Responsible open sourcing is linked to greater transparency across the entire the model pipeline. 
As illustrated by Table \ref{tab:classification_full} (Appendix \ref{app:taxonomy_tables}), a lack of data transparency is a problem even in relatively open LLMs. Making training and evaluation data publicly available enhances the community's capacity to scrutinize models' capabilities, risks, and limitations, thereby unlocking many of the advantages outlined in \S \ref{sec:near-to-mid-risks}. 
It also holds the potential to develop models pre-trained for safety rather than aligned post-hoc. We believe this is an area where more research is needed which requires more parts of the pipeline to be open.
Additionally, transparency in dataset composition, including metadata like copyright, is crucial. Maintaining comprehensive audit logs detailing chains of custody, transformations, data augmentation, and synthesis processes is increasingly vital.

\textbf{Improve Open Evaluation and Benchmarking }
There has been much progress in open benchmarking of general LLM capabilities (e.g. LMSys, HELM, AlpacaEval), but there is an outstanding need for benchmarks that are specific to particular domains and impact areas, including model safety. This is poignant since, as highlighted in \S \ref{sec:taxonomy}, most developers do not release their safety training and evaluation data. Generally, new models should be evaluated pre-release, so that their capabilities, risks, and limitations are made clear from day one. Evaluations should include assessments as related to the variety of risks outlined in \S \ref{sec:near-to-mid-risks}.

\textbf{Conduct Multilevel Security Audits } 
Open source affords pre- and fine-tuning of models for any downstream tasks. For mission-critical tasks, particularly in areas like mental health, multi-level security audits and procedures must be meticulously designed, documented, implemented, and publicly reported. This should encompass both manual and automated testing, ranging from adversarial jailbreak prompts to expert-led red-teaming for common and edge case exploits, where financially viable. Additionally, incorporating static and dynamic analysis toolchains into developers' IDEs is essential to detect vulnerabilities early in the development process. Establishing and promoting safe design patterns for Gen AI development within the community is also crucial. Once ready for deployment, it is important that developers engage with the wider safety research community to allow for further third-party testing in controlled sandboxes closer to the released model environment.

\textbf{Compare with Closed Source Models }
Open-source models offer advantages like enhanced privacy, customization, transparency, efficiency, and cost-effectiveness. In contrast, commercial closed-source models can stand out in performance, usability, and liability protections. Therefore, comparing the models with their closest commercial closed source alternative is important to quantify, clarify, and understand the trade-offs involved in open sourcing decisions.

\textbf{Conduct Studies of Broader Societal Impact }
As highlighted in \S \ref{sec:societal_impact}, properly developed open models can reduce Gen AI energy consumption, aid in resolving copyright disputes, cater to diverse communities, and help democratize AI development. To realize these benefits, it's crucial to undertake comprehensive broader societal impact studies. These should include evaluating corporate practices in model design and management, initiatives for enhancing data diversity and representation, and transparency reports on the environmental impact of the models.

\section{Conclusion}
\label{sec:conclusion}
The recommendations in \S \ref{sec:recs} are a result of combining the openness trends of currently available models in \S \ref{sec:taxonomy} with the analysis of \S \ref{sec:near-to-mid-risks} on the potential risks and opportunities of open sourcing near to mid-term models. 
Following this discussion, we advocate for the \textbf{responsible open sourcing of near to mid-term Gen AI models}. 

Note that our position is a balanced one. We advocate that developers should be allowed and encouraged to responsibly open-source Gen AI models developed in the near to mid-term stages, in as much as it makes economic sense for them to do so. Building Gen AI models is an expensive process, and we are sensitive to the argument that for-profit companies should be able to reap some of the financial benefits of their investments in building the technology. Any other position on this matter (e.g., forcing companies to open source their models/pipelines) would seriously risk investment and progress in this area.

However, often for-profit entities will claim open source Gen AI is fundamentally unsafe, and will publicly use this to argue against the open sourcing of these models altogether. This discourages other developers from open sourcing, and we believe this is one of the main factors that contributes to the current skew in the landscape presented in the taxonomy of \S \ref{sec:taxonomy} (Figure \ref{fig:classification_distribution}). We reject this argument, and argue in \S \ref{sec:near-to-mid-risks} and \S \ref{sec:recommendations} that (1) there are many benefits that can only be achieved through open sourcing, and (2) the risks are often exaggerated by these for-profit entities. By making these impacts explicit and laying out recommendations for the responsible open sourcing of these models, our aim is to encourage developers to improve the notable skew in Figure \ref{fig:classification_distribution}. This does not mean all models will be open-sourced, only that there would be an improved balance. We note that this should always be voluntary rather than imposed, to avoid disrupting the investment in the area.

Our work underscores the importance of mitigating risks and addresses prevalent concerns, thereby paving the way for realizing the vast potential benefits open-source generative AI offers.

\section{Related Work}
The debate around open sourcing GenAI differs from the well-studied impacts of open-source software on society \citep{jaisingh2008impact} due to the unique characteristics of the technology.
As such, we report related works on two axes: (1) examining the broader impact of GenAI, and (2) on the debate around open sourcing these models.\looseness=-1


\textbf{The Impact of GenAI }
There are many works that focus on the risks and benefits of the technology as it exists today, particularly with respect to areas such as science \& medicine~\cite{ai4science2023impact,fecher2023friend}, education~\cite{alahdab2023potential,cooper2023examining,malik2023exploring}, the environment~\cite{rillig2023risks}, among others. Other research evaluates the potential impacts of a capability shift \citet{seger2023open}, emphasizing the critical importance of transparency in analyzing AI failures \citep{kapoor_narayanan_2023, kapoor_narayanan_2023b}. 

\textbf{On Open Sourcing GenAI Models }
A main line of discussion centers on the definition of open sourcing GenAI, highlighting the role of disclosing the training pipeline, weights, and data in achieving the benefits of open source \citep{bommasani2023foundation,bommasani2,liesenfeld2023opening,seger2023open,shrestha2023building}. Notably, AI systems typically encompass more than just code, necessitating custom release pipelines~\cite{liu2023llm360}. Others~\cite{laion, hacker2023regulating, tumadottir} highlight the need to differentiate open-source systems from a regulatory standpoint, to avoid compliance costs unsustainable for open source contributors~\cite{euact}. 
Many highlight the risks of centralization in absence of open source~\cite{seger2023open,horowitz}. On the other hand, open models may exacerbate the risks of misuse~\cite{bommasani2021opportunities,alaga2023coordinated} unless proper measures are instituted for responsibly open-sourcing them. Interestingly, it has also been shown that open GenAI tends to be less trustworthy than closed ones~\cite{sun2024trustllm}. A relevant paper~\cite{seger2023open} analyzes the risks and benefits of open models, and shapes recommendations for the near future. In our work, we provide a holistic viewpoint centered on near to mid-term models, including a taxonomy of the current landscape and discussion of future impacts.

\section*{Impact Statement}
\vspace{-0.1cm}
This work presents an attempt at a comprehensive evaluation of the risks and benefits associated with open-sourcing generative AI models as well as a list of prescriptions for responsible open-sourcing. The speculative nature of our work comes naturally with a broad impact potential. From a regulatory viewpoint, this paper could influence policy makers in the decision-making process concerning lawmaking oriented to open-source generative AI. Also, the impact on companies and open-source communities' release processes is potentially significant, considering the recent extremely high interest in developing and releasing open-source models. We stress that although our analysis is thorough, our risk assessment has fundamental assumptions that must be respected, and re-evaluated in case of disruptive unpredictable changes violating our hypotheses.

\section*{Disclaimer}
\vspace{-0.1cm}
This paper represents the collaborative effort of a diverse group of researchers, each bringing their own unique perspectives to the table. We note that not every viewpoint expressed within this work is necessarily unanimously agreed upon by all authors.

\section*{Acknowledgments}
\vspace{-0.1cm}
The authors would like to thank Meta for their generous support, including travel grants and logistical assistance, which enabled this collaboration, as well as for the organization of the first Open Innovation AI Research Community workshop where this work was initiated. Meta had no editorial input in this paper, and the views expressed herein do not reflect those of the company.

FE is supported by EPSRC Centre for Doctoral Training in
Autonomous Intelligent Machines and Systems [EP/S024050/1] and Five AI Limited.
AP is funded by EPSRC Centre for Doctoral Training in
Autonomous Intelligent Machines and Systems [EP/S024050/1].
FP is funded by KAUST (Grant DFR07910).
JMI is funded by National University Philippines and the UKRI Centre for Doctoral Training in Accountable, Responsible and Transparent AI [EP/S023437/1] of the University of Bath.
PR is supported by a MUR FARE 2020 initiative under grant agreement Prot. R20YSMBZ8S (INDOMITA).
PHST is supported by UKRI grant: Turing AI Fellowship EP/W002981/1, and by the Royal Academy of Engineering under the Research Chair and Senior Research Fellowships scheme.
JF is partially funded by the UKI grant EP/Y028481/1 (originally selected for funding by the ERC).

\bibliography{refs}
\bibliographystyle{acl_natbib}

\appendix
\onecolumn

\onecolumn

\section{Further details on training, evaluation and deployment}
\label{app:pipeline_details}

Model training (1) processes can be grouped into three distinct stages:
\begin{enumerate}
    \item \textit{Pre-training}, where a model is exposed to large-scale datasets composed of trillions of tokens of data, typically scraped from the internet and usually uncurated. The goal is for the model to see a diversity of data, and through that process develop fundamental skills (e.g., grammar, vocabulary, text structure) and broad knowledge \citep{gao2020pile,radford2019language}. An example of a commonly used open source dataset for pre-training LLMs such as LLaMA or GPT-J is The Pile which combines 22 smaller datasets into a diverse 825Gb text dataset \cite{gao2020pile,touvron2023llama,gpt-j}.
    \item \textit{Supervised fine-tuning (SFT)}, which is intended to correct for data quality issues in pre-training datasets. Usually, a much smaller amount of high quality data is used to improve model performance. Several works observe that at this stage the quality of the data used is essential to the downstream performance of the models \citep{zhou2024lima, ouyang2022training, touvron2023llama2, team2023gemini}, with the authors of LLaMA-2 pointing out that \textit{``by setting aside millions of examples from third-party datasets and using fewer but higher-quality examples from our own vendor-based annotation efforts, [their] results notably improved.''} \citep{touvron2023llama2}.
    \item \textit{Alignment}, which is used to create an application-specific version of the foundation model (e.g., a chatbot or translation model). Reinforcement Learning with Human Feedback (RLHF) or Direct Preference Optimisation (DPO) \citep{ouyang2022training, touvron2023llama2} is used to create a model that follows instructions and is better-aligned with human preferences. With RLHF, a dataset of human preferences over model outputs is used to train a Reward model, which in turn is used with a reinforcement learning algorithm (e.g., PPO; \citeauthor{schulman2017proximal}, \citeyear{schulman2017proximal}) to align the LLM. RLHF is not used in models released prior to 2022 \citep{brown2020language, xue2020mt5, smith2022using}, and it is unclear whether the RLHF is used in models such as PaLM-2 \citep{anil2023palm}.
\end{enumerate}

Once trained, models are usually evaluated (2) on openly available evaluation datasets such as MMLU or NaturalQuestions \citep{hendrycks2020measuring, kwiatkowski2019natural} as well as curated benchmarks such as HELM, BigBench EleutherAI’s Evaluation Harness \citep{liang2022holistic, srivastava2022beyond, eval-harness}. Some models are also evaluated on proprietary datasets held internally by developers, potentially by holding out some of the SFT/RLHF data from the training process \citep{touvron2023llama2}. However, there is little publicly available information on how this is implemented, and few details are shared about the composition of such datasets. On top of utility-based benchmarking, developers sometimes create safety evaluation mechanisms to proactively stress-test the outputs of the model. These include human-annotated safety evaluation datasets (e.g., through creating adversarial prompts), as well as automatic safety evaluation algorithms \citep{touvron2023llama2, yuan2023gpt4}. They are typically the result of applying techniques such as red teaming. Finally, at the deployment stage (3), content can be generated by running the inference code with the associated model weights.

\section{Full Taxonomy Tables}
\label{app:taxonomy_tables}

\textbf{Important disclaimer:} Table \ref{tab:classification_full} focuses on component openness in model pipelines, not reproducibility. GLM-130B and Falcon provide detailed training procedures, unlike GPT-4, yet those are all classified as C1 due to unreleased pre-training code. A full reproducibility assessment falls beyond this report's scope. 

\begin{table*}
    \begin{adjustwidth}{0pt}{0pt}
    {
    \scriptsize
    \centering
    \begin{tabular}{>{\hspace{0pt}}m{0.10\linewidth}>{\centering\hspace{0pt}}m{0.07\linewidth}>{\centering\hspace{0pt}}m{0.09\linewidth}>{\centering\hspace{0pt}}m{0.09\linewidth}>{\centering\hspace{0pt}}m{0.09\linewidth}>{\centering\hspace{0pt}}m{0.09\linewidth}>{\centering\hspace{0pt}}m{0.11\linewidth}>{\centering\hspace{0pt}}m{0.07\linewidth}>{\centering\arraybackslash\hspace{0pt}}m{0.07\linewidth}}
    \toprule
    \textbf{License}        & \textbf{Research} & \textbf{Commercial Purposes} & \textbf{Modify as Desired} & \textbf{Copyright derivative work} & \textbf{Other license for derivative} & \textbf{Final score}            & \textbf{Code Openness} & \textbf{Data Openness}  \\ \midrulenospacing
    MIT/Mod. MIT            & Y                 & Y                            & Y                          & Y                                  & Y                                     & 5\par{}(Restriction free)       & \CFiveTable                     & \DFiveTable                      \\ \hline
    Apache 2.0              & Y                 & Y                            & Y                          & Y                                  & Y                                     & 5\par{}(Restriction free)       & \CFiveTable                     & \DFiveTable                      \\ \hline
    Common Crawl\par{}(ComCrawl) & Y                 & Y                            & Y                          & Y                                  & Y                                     & 5\par{}(Restriction free)       & \CFiveTable                     & \DFiveTable                      \\ \hline
    BSD-3                   & Y                 & Y                            & Y                          & Y                                  & Y                                     & 5\par{}(Restriction free)       & \CFiveTable                     & \DFiveTable                      \\ \hline
    RAIL                    & Y                 & Y                            & Y                          & Y                                  & N                                     & 4\par{}(Slightly restrictive)   & \CFourTable                     & \DFourTable                      \\ \hline
    LLaMA-2                 & Y                 & Y\footnote{For models with up to 700M users.}                            & N                          & Y                                  & N                                     & 3\par{}(Moderately restrictive) & \CThreeTable                     & \DThreeTable                      \\ \hline
    ODC-By                  & Y                 & Y                            & Y                          & Y                                  & N                                     & 4\par{}(Slightly restrictive)   & \NATable                    & \DFourTable                      \\ \hline
    CodeT5 Data             & Y                 & Y                            & Y                          & Y                                  & N                                     & 4\par{}(Slightly restrictive)   & \NATable                    & \DFourTable                      \\ \hline
    RedPajama Data\par{}(Full)   & Y                 & Y                            & Y                          & Y                                  & N                                     & 4\par{}(Slightly restrictive)   & \NATable                    & \DFourTable                      \\ \hline
    OPT Data                & Y                 & N                            & N                          & N                                  & N                                     & 1\par{}(Highly restrictive)     & \NATable                    & \DThreeTable                      \\ \hline
    GLM-130B Data           & Y                 & N                            & N                          & N                                  & N                                     & 1\par{}(Highly restrictive)     & \NATable                    & \DThreeTable                      \\ \hline
    Falcon-180B Data        & Y                 & Y                            & Y                          & Y                                  & Y                                     & 5\par{}(Restriction free)       & \NATable                    & \DFiveTable               \\ \bottomrulenospacing
    \end{tabular}
    }
    \vspace{0.5em}
    \caption{\textbf{License Openness Taxonomy}: categorization of commonly used licenses in a variety of relevant open source criteria, and resulting code and data openness categories.}
    \label{tab:licenses_full}
    \end{adjustwidth}
\end{table*}

\renewcommand{\arraystretch}{1.2}
\begin{table*}
    {
    \scriptsize
    \centering

    \begin{tabular}{>{\hspace{0pt}}m{0.11\linewidth}>{\centering\hspace{0pt}}m{0.12\linewidth}>{\centering\hspace{0pt}}m{0.08\linewidth}>{\centering\hspace{0pt}}m{0.09\linewidth}>{\centering\hspace{0pt}}m{0.12\linewidth}>{\centering\hspace{0pt}}m{0.13\linewidth}>{\centering\hspace{0pt}}m{0.15\linewidth}>{\hspace{0pt}}m{0.00001\linewidth}}

    \toprule
    \multirow{2}[6]{\linewidth}{\hspace{0pt}\textbf{Model}} & \multirow{2}[6]{\linewidth}{\hspace{0pt}\Centering{}\textbf{Developer}} & \multirow{2}[6]{\linewidth}{\hspace{0pt}\Centering{}\textbf{Largest Model Size (params)}} & \multirow{2}[6]{\linewidth}{\hspace{0pt}\Centering{}\textbf{Release Date}} & \multicolumn{3}{>{\Centering\hspace{0pt}}m{0.41\linewidth}}{\textbf{Impact Metrics}} & \\ \cmidrule{5-7}
    
    & & & & \textbf{ChatBot Arena Elo Rating} & \textbf{Google Scholar Citations} & \textbf{HuggingFace Downloads Last Month} & \\ \midrulenospacing
    GPT-2                                                     & \DevCompany{OpenAI}                                                                    & 1.5B                                                                                       & 02/2019                                                                      & N/A                               & 8,015                             & 17,984,300                                & \\ \hline
    T5                                                        & \DevCompany{Google}                                                                    & 11B                                                                                        & 10/2019                                                                      & 873                               & 12,162                            & 3,295,844                                 & \\ \hline
    GPT-3                                                     & \DevCompany{OpenAI}                                                                    & 175B                                                                                       & 05/2020                                                                      & N/A                               & 18,759                            & N/A                                       & \\ \hline
    mT5                                                       & \DevCompany{Google}                                                                    & 13B                                                                                        & 10/2020                                                                      & N/A                               & 1,439                             & 631,429                                   & \\ \hline
    GPT-Neo                                                   & \DevNonProfit{EleutherAI}                                                                & 2.7B                                                                                       & 03/2021                                                                      & N/A                               & N/A                               & 242,580                                   & \\ \hline
    GPT-J-6B                                                  & \DevNonProfit{EleutherAI}                                                                & 6B                                                                                         & 06/2021                                                                      & N/A                               & 465                               & 95,620                                    & \\ \hline
    CodeT5                                                    & \DevCompany{Salesforce}                                                                & 16B                                                                                        & 09/2021                                                                      & N/A                               & 703                               & 23,549                                    & \\ \hline
    Megatron-Turing                                           & \DevCompany{Microsoft, NVIDIA}                                                         & 530B                                                                                       & 10/2021                                                                      & N/A                               & 379                               & N/A                                       & \\ \hline
    Anthropic LM                                              & \DevCompany{Anthropic}                                                                 & 52B                                                                                        & 12/2021                                                                      & N/A                               & 70                                & N/A                                       & \\ \hline
    ERNIE 3.0                                                 & \DevCompany{Baidu}                                                                     & 260B                                                                                       & 12/2021                                                                      & N/A                               & 248                               & 728                                       & \\ \hline
    Gopher                                                    & \DevCompany{DeepMind}                                                                  & 280B                                                                                       & 12/2021                                                                      & N/A                               & 598                               & N/A                                       & \\ \hline
    GLaM                                                      & \DevCompany{Google}                                                                    & 1.2T                                                                                       & 12/2021                                                                      & N/A                               & 255                               & N/A                                       & \\ \hline
    XGLM                                                      & \DevCompany{Meta}                                                                      & 7.5B                                                                                       & 12/2021                                                                      & N/A                               & 79                                & 12,884                                    & \\ \hline
    FairSeq Dense                                             & \DevCompany{Meta}                                                                      & 13B                                                                                        & 12/2021                                                                      & N/A                               & 34                                & 6,129                                     & \\ \hline
    LaMDA                                                     & \DevCompany{Google}                                                                    & 127B                                                                                       & 01/2022                                                                      & N/A                               & 819                               & N/A                                       & \\ \hline
    GPT-NeoX-20B                                              & \DevNonProfit{EleutherAI}                                                                & 20B                                                                                        & 02/2022                                                                      & N/A                               & 364                               & 37,122                                    & \\ \hline
    PolyCoder                                                 & \DevNonProfit{Carnegie Mellon}                                                           & 2.7B                                                                                       & 02/2022                                                                      & N/A                               & 259                               & 554                                       & \\ \hline
    Chinchilla                                                & \DevCompany{DeepMind}                                                                  & 70B                                                                                        & 03/2022                                                                      & N/A                               & 245                               & N/A                                       & \\ \hline
    PaLM                                                      & \DevCompany{Google}                                                                    & 540B                                                                                       & 04/2022                                                                      & 1,004                             & 2,342                             & N/A                                       & \\ \hline
    OPT                                                       & \DevCompany{Meta}                                                                      & 175B                                                                                       & 05/2022                                                                      & N/A                               & 1,105                             & 191,115                                   & \\ \hline
    UL2                                                       & \DevCompany{Google}                                                                    & 20B                                                                                        & 05/2022                                                                      & N/A                               & 99                                & 20,731                                    & \\ \hline
    BLOOM                                                     & \DevNonProfit{Big Science}                                                               & 176B                                                                                       & 05/2022                                                                      & N/A                               & 814                               & 1,172,142                                 & \\ \hline
    GLM-130B                                                  & \DevNonProfit{Tsinghua University}                                                       & 130B                                                                                       & 10/2022                                                                      & N/A                               & 129                               & 345                                       & \\ \hline
    Pythia                                                    & \DevNonProfit{EleutherAI}                                                                & 12B                                                                                        & 12/2022                                                                      & 896                               & 195                               & 55,398                                    & \\ \hline
    Anthropic 175B LM                                         & \DevCompany{Anthropic}                                                                 & 175B                                                                                       & 02/2023                                                                      & N/A                               & 55                                & N/A                                       & \\ \hline
    LLaMA                                                     & \DevCompany{Meta}                                                                      & 13B                                                                                        & 02/2023                                                                      & 800                               & 2,793                             & N/A                                       & \\ \hline
    GPT-4                                                     & \DevCompany{OpenAI}                                                                    & N/A                                                                                        & 03/2023                                                                      & 1,243                             & 308                               & N/A                                       & \\ \hline
    Claude                                                    & \DevCompany{Anthropic}                                                                 & N/A                                                                                        & 03/2023                                                                      & 1,149                             & N/A                               & N/A                                       & \\ \hline
    Cerebras-GPT                                              & \DevCompany{Cerebras}                                                                  & 13B                                                                                        & 03/2023                                                                      & N/A                               & 23                                & 124,561                                   & \\ \hline
    Stable LM                                                 & \DevCompany{Stability AI}                                                              & 7B                                                                                         & 04/2023                                                                      & 844                               & N/A                               & 15,282                                    & \\ \hline
    PaLM-2                                                    & \DevCompany{Google}                                                                    & N/A                                                                                        & 05/2023                                                                      & N/A                               & 372                               & N/A                                       & \\ \hline
    OpenLLaMA                                                 & \DevNonProfit{UC Berkeley}                                                               & 13B                                                                                        & 06/2023                                                                      & N/A                               & N/A                               & 58,991                                    & \\ \hline
    Claude-2                                                  & \DevCompany{Anthropic}                                                                 & N/A                                                                                        & 07/2023                                                                      & 1,131                             & N/A                               & N/A                                       & \\ \hline
    LLaMA-2                                                   & \DevCompany{Meta}                                                                      & 70B                                                                                        & 07/2023                                                                      & 1,077                             & 1,197                             & 742,238                                   & \\ \hline
    Falcon                                                    & \DevGovernment{TII}                                                                       & 180B                                                                                       & 09/2023                                                                      & 1,035                             & 65                                & 1,341,297                                 & \\ \hline
    GPT-3.5-turbo                                             & \DevCompany{OpenAI}                                                                    & N/A                                                                                        & 09/2023                                                                      & 1,117                             & N/A                               & N/A                                       & \\ \hline
    Mistral-7B                                                & \DevCompany{Mistral AI}                                                                & 7B                                                                                         & 10/2023                                                                      & 1,023                             & 15                                & 510,471                                   & \\ \hline
    Grok-1                                                    & \DevCompany{xAI}                                                                       & N/A                                                                                        & 11/2023                                                                      & N/A                               & N/A                               & N/A                                       & \\ \hline
    Phi-2                                                     & \DevCompany{Microsoft}                                                                 & 2.7B                                                                                       & 11/2023                                                                      & N/A                               & N/A                               & 85,200                                    & \\ \hline
    Gemini                                                    & \DevCompany{Google DeepMind}                                                           & N/A                                                                                        & 12/2023                                                                      & 1,111                             & N/A                               & N/A                                       &
   \\ \bottomrulenospacing
    
    \end{tabular}
    }
    \vspace{0.5em}
    \caption{\textbf{Model Information}: table containing the basic information about each of the models classified under the openness taxonomy. Developers highlighted in \colorbox{CompanyColor}{purple} correspond to companies, in \colorbox{NonProfitColor}{pink} are non-profit entities, and in \colorbox{GovernmentColor}{light blue} are government institutes. All data accessed on 28th of December 2023.}
    \label{tab:model_list_full}
\end{table*}

\clearpage
\onecolumn

{
    \scriptsize
    \begin{longtable}{
        >{\hspace{0pt}}m{0.06\linewidth}|
        >{\centering\hspace{0pt}}m{0.04\linewidth}>{\centering\hspace{0pt}}m{0.04\linewidth}>{\centering\hspace{0pt}}m{0.04\linewidth}|
        >{\centering\hspace{0pt}}m{0.04\linewidth}>{\centering\hspace{0pt}}m{0.05\linewidth}>{\centering\hspace{0pt}}m{0.04\linewidth}|
        >{\centering\hspace{0pt}}m{0.05\linewidth}>{\centering\hspace{0pt}}m{0.07\linewidth}|
        >{\centering\hspace{0pt}}m{0.05\linewidth}>{\centering\hspace{0pt}}m{0.05\linewidth}|
        >{\centering\hspace{0pt}}m{0.06\linewidth}|>{\centering\arraybackslash\hspace{0pt}}m{0.08\linewidth}
    }
    
    \toprule
    \multirow{3}[6]{0.06\linewidth}{\hspace{0pt}\textbf{Model}} & \multicolumn{6}{c}{\textbf{(1) Training}} & \multicolumn{4}{c}{\textbf{(2) Evaluation}}                                                                      & \multicolumn{2}{c}{\textbf{(3) Deployment}}  \\ \cmidrule{2-13}
    
    & \multicolumn{3}{c}{\textbf{Code}} & \multicolumn{3}{c}{\textbf{Data}} & \multicolumn{2}{c}{\textbf{Code}} & \multicolumn{2}{c}{\textbf{Data}} & \multicolumn{1}{c}{\textbf{Code}}          & \textbf{Data}                                                                \\ \cmidrule{2-13} 
    
    & \textbf{Pre-Training}  & \textbf{Fine-tuning}   & \textbf{Alignment} & \textbf{Pre-Training}                         & \textbf{Supervised FT} & \textbf{Alignment} & \textbf{General Eval}  & \textbf{Automatic Safety Eval}         & \textbf{Utility Benchmarks} & \textbf{Safety Eval Datasets}   & \textbf{Inference}     & \textbf{Model Architecture and Weights}                                      \endfirsthead
    \midrulenospacing
GPT-2                                     & \COneTable                 & \NATable                & \NATable & \DOneTable                     & \NATable                & \NATable         & \COneTable                         & \NATable & \DOneTable  & \NATable                        & {\CFiveTable\par{}(Mod. MIT)}     & {\DFiveTable\par{}(Mod. MIT)}         \\\hline
T5                                        & {\CFiveTable\par{}(Apache 2.0)} & {\CFiveTable\par{}(Apache 2.0)} & \NATable & {\DFourTable\par{}(ODC-By)}         & \NATable                & \NATable         & {\CFiveTable\par{}(Apache 2.0)}         & \NATable & \NATable & \NATable                        & {\CFiveTable\par{}(Apache 2.0)}   & {\DFiveTable\par{}(Apache 2.0)}       \\\hline
GPT-3                                     & \COneTable                 & \COneTable                 & \NATable & \DOneTable                     & \NATable                & \NATable         & \COneTable                         & \NATable & \DOneTable  & \NATable                        & \COneTable                   & \DTwoTable                       \\\hline
mT5                                       & {\CFiveTable\par{}(Apache 2.0)} & {\CFiveTable\par{}(Apache 2.0)} & \NATable & {\DFourTable\par{}(ODC-By)}         & \NATable                & \NATable         & {\CFiveTable\par{}(Apache 2.0)}         & \NATable & \NATable & \NATable                        & {\CFiveTable\par{}(Apache 2.0)}   & {\DFiveTable\par{}(Apache 2.0)}       \\\hline
GPT-Neo                                   & {\CFiveTable\par{}(MIT)}        & {\CFiveTable\par{}(MIT)}        & \NATable & {\DFiveTable\par{}(MIT)}            & \NATable                & \NATable         & {\CFiveTable\par{}(MIT)}                & \NATable & \NATable & \NATable                        & {\CFiveTable\par{}(MIT)}          & {\DFiveTable\par{}(MIT)}              \\\hline
GPT-J-6B                                  & {\CFiveTable\par{}(Apache 2.0)} & {\CFiveTable\par{}(Apache 2.0)} & \NATable & {\DFiveTable\par{}(MIT)}            & \NATable                & \NATable         & {\CFiveTable\par{}(Apache 2.0)}         & \NATable & \NATable & \NATable                        & {\CFiveTable\par{}(Apache 2.0)}   & {\DFiveTable\par{}(Apache 2.0)}       \\\hline
CodeT5                                    & {\CFiveTable\par{}(BSD-3)}      & {\CFiveTable\par{}(BSD-3)}      & \NATable & {\DFourTable\par{}(CodeT5)}         & \NATable                & \NATable         & {\CFiveTable\par{}(BSD-3)}              & \NATable & \NATable & \NATable                        & {\CFiveTable\par{}(BSD-3)}        & {\DFiveTable\par{}(Apache 2.0)}       \\\hline
Megatron-Turing                           & \COneTable                 & \NATable                & \NATable & \DOneTable                     & \NATable                & \NATable         & \COneTable                         & \NATable & \NATable & \NATable                        & \COneTable                   & \DOneTable                       \\\hline
Anthropic LM                              & \COneTable                 & \COneTable                 & \NATable & \DOneTable                     & \NATable                & {\DFiveTable\par{}(MIT)} & \COneTable                         & \NATable & \NATable & {\DFiveTable\par{}(MIT)}                & \COneTable                   & \DOneTable                       \\\hline
ERNIE 3.0                                 & \COneTable                 & \COneTable                 & \NATable & \DOneTable                     & \NATable                & \NATable         & \COneTable                         & \NATable & \NATable & \NATable                        & \COneTable                   & \DOneTable                       \\\hline
Gopher                                    & \COneTable                 & \COneTable                 & \NATable & \DOneTable                     & \NATable                & \NATable         & \COneTable                         & \NATable & \DOneTable  & \DOneTable                         & \COneTable                   & \DOneTable                       \\\hline
GLaM                                      & \COneTable                 & \NATable                & \NATable & \DOneTable                     & \NATable                & \NATable         & \COneTable                         & \NATable & \NATable & \NATable                        & \COneTable                   & \DOneTable                       \\\hline
XGLM                                      & {\CFiveTable\par{}(MIT)}        & \NATable                & \NATable & {\DFiveTable\par{}(ComCrawl)}       & \NATable                & \NATable         & {\CFiveTable\par{}(MIT)}                & \COneTable  & \NATable & {\DFiveTable\par{}(Public datasets)}    & {\CFiveTable\par{}(MIT)}          & {\DFiveTable\par{}(MIT)}              \\\hline
FairSeq Dense                             & {\CFiveTable\par{}(MIT)}        & \NATable                & \NATable & {\DFiveTable\par{}(ComCrawl)}       & \NATable                & \NATable         & \NATable                        & \NATable & \NATable & \NATable                        & {\CFiveTable\par{}(MIT)}          & {\DFiveTable\par{}(MIT)}              \\\hline
LaMDA                                     & \COneTable                 & \COneTable                 & \NATable & \DOneTable                     & \DOneTable                 & \NATable         & \COneTable                         & \COneTable  & \DOneTable  & \DOneTable                         & \COneTable                   & \DOneTable                       \\\hline
GPT-NeoX-20B                              & {\CFiveTable\par{}(Apache 2.0)} & \NATable                & \NATable & {\DFiveTable\par{}(MIT)}            & \NATable                & \NATable         & {\CFiveTable\par{}(Apache 2.0)}         & \NATable & \NATable & \NATable                        & {\CFiveTable\par{}(Apache 2.0)}   & {\DFiveTable\par{}(Apache 2.0)}       \\\hline
PolyCoder                                 & {\CFiveTable\par{}(MIT)}        & \NATable                & \NATable & {\UnkTable~\par{}(D3 or D4)}       & \NATable                & \NATable         & {\CFiveTable\par{}(MIT)}                & \NATable & \NATable & \NATable                        & {\CFiveTable\par{}(CC BY-SA-4.0)} & {\DFiveTable\par{}(CC BY-SA-4.0)}     \\\hline
Chinchilla                                & \COneTable                 & \COneTable                 & \NATable & \DOneTable                     & \NATable                & \NATable         & \COneTable                         & \NATable & \NATable & \NATable                        & \COneTable                   & \DOneTable                       \\\hline
PaLM                                      & \COneTable                 & \COneTable                 & \NATable & \DOneTable                     & \DOneTable                 & \NATable         & \COneTable                         & \NATable & \NATable & \NATable                        & \COneTable                   & \DOneTable                       \\\hline
OPT                                       & {\CFiveTable\par{}(MIT)}        & \NATable                & \NATable & \UnkTable                      & \NATable                & \NATable         & \COneTable                         & \NATable & \NATable & \NATable                        & {\CFiveTable\par{}(MIT)}          & {\DThreeTable\par{}(OPT Data)}         \\\hline
UL2                                       & {\CFiveTable\par{}(Apache 2.0)} & {\CFiveTable\par{}(Apache 2.0)} & \NATable & {\DFourTable\par{}(ODC-By)}         & \NATable                & \NATable         & {\CFiveTable\par{}(Apache 2.0)}         & \NATable & \NATable & \NATable                        & {\CFiveTable\par{}(Apache 2.0)}   & {\DFiveTable\par{}(Apache 2.0)}       \\\hline
BLOOM                                     & {\CFiveTable\par{}(Apache 2.0)} & \UnkTable                  & \NATable & {\UnkTable~\par{}(D3 or D4)}       & {\DFiveTable\par{}(Apache 2.0)} & \NATable         & {\CFiveTable\par{}(Apache 2.0)}         & \NATable & \NATable & \NATable                        & {\CFiveTable\par{}(Apache 2.0)}   & {\DFourTable\par{}(RAIL)}             \\\hline
GLM-130B                                  & \COneTable                 & \NATable                & \NATable & \DOneTable                     & \NATable                & \NATable         & {\CFiveTable\par{}(Apache 2.0)}         & \NATable & \NATable & \NATable                        & {\CFiveTable\par{}(Apache 2.0)}   & {\DThreeTable\par{}(GLM-130B Data)}    \\\hline
Pythia                                    & {\CFiveTable\par{}(Apache 2.0)} & \NATable                & \NATable & {\DFiveTable\par{}(MIT)}            & \NATable                & \NATable         & {\CFiveTable\par{}(Apache 2.0)}         & \NATable & \NATable & \NATable                        & {\CFiveTable\par{}(Apache 2.0)}   & {\DFiveTable\par{}(Apache 2.0)}       \\\hline
Anthropic 175B                            & \COneTable                 & \COneTable                 & \COneTable  & \DOneTable                     & \DOneTable                 & \DOneTable          & \COneTable                         & \NATable & \NATable & \DOneTable                         & \COneTable                   & \DOneTable                       \\\hline
LLaMA                                     & \COneTable                 & \NATable                & \NATable & {\UnkTable~\par{}(likely D5)}      & \NATable                & \NATable         & \COneTable                         & \COneTable  & \NATable & {\DFiveTable\par{}(Publicly available)} & {\CFourTable\par{}(GNU GPL)}      & {\DThreeTable\par{}(LLaMA)}            \\\hline
GPT-4                                     & \COneTable                 & \COneTable                 & \COneTable  & \DOneTable                     & \DOneTable                 & \DOneTable          & {\CFiveTable\par{}(MIT)}                & \NATable & \DOneTable  & \DOneTable                         & \COneTable                   & \DTwoTable                       \\\hline
Claude                                    & \COneTable                 & \COneTable                 & \COneTable  & \DOneTable                     & \DOneTable                 & \DOneTable          & \COneTable                         & \NATable & \NATable & \DOneTable                         & \COneTable                   & \DOneTable                       \\\hline
Cerebras-GPT                              & {\CFiveTable\par{}(Apache 2.0)} & \NATable                & \NATable & {\DFiveTable\par{}(MIT)}            & \NATable                & \NATable         & {\CFiveTable\par{}(Publicly available)} & \NATable & \NATable & \NATable                        & {\CFiveTable\par{}(Apache 2.0)}   & {\DFiveTable\par{}(Apache 2.0)}       \\\hline
Stable LM & \COneTable                 & \COneTable                 & \NATable & {\DFourTable\par{}(CC BY-SA-4.0)}   & \DOneTable                 & \NATable         & \COneTable                         & \NATable & \NATable & \NATable                        & {\CFiveTable\par{}(CC BY-SA-4.0)} & {\DFiveTable\par{}(CC BY-SA-4.0)}     \\\hline
PaLM-2                                    & \COneTable                 & \NATable                & \NATable & \DOneTable                     & \NATable                & \NATable         & \COneTable                         & \NATable & \NATable & {\DFiveTable\par{}(Publicly available)} & \COneTable                   & \DOneTable                       \\\hline
OpenLLaMA                                 & {\CFiveTable\par{}(Apache 2.0)} & \NATable                & \NATable & {\DFourTable\par{}(RedPajama Data)} & \NATable                & \NATable         & {\CFiveTable\par{}(Apache 2.0)}         & \NATable & \NATable & \NATable                        & {\CFiveTable\par{}(Apache 2.0)}   & {\DFiveTable\par{}(Apache 2.0)}       \\\hline
Claude-2                                  & \COneTable                 & \COneTable                 & \COneTable  & \DOneTable                     & \DOneTable                 & \DOneTable          & \COneTable                         & \COneTable  & \DOneTable  & \DOneTable                         & \COneTable                   & \DTwoTable                       \\\hline
LLaMA-2                                   & \COneTable                 & \COneTable                 & \COneTable  & \DOneTable                     & \DOneTable                 & \DOneTable          & \COneTable                         & \NATable & \NATable & \DOneTable                         & {\CThreeTable\par{}(LLaMA-2)}      & {\DThreeTable\par{}(LLaMA-2)}          \\\hline
Falcon                                    & \COneTable                 & \COneTable                 & \COneTable  & {\DFourTable\par{}(ODC-By)}         & \DOneTable                 & \DOneTable          & \COneTable                         & \NATable & \NATable & \NATable                        & {\CFiveTable\par{}(Apache 2.0)}   & {\DFiveTable\par{}(Falcon-180B Data)} \\\hline
GPT-3.5-turbo                             & \COneTable                 & \COneTable                 & \COneTable  & \DOneTable                     & \DOneTable                 & \DOneTable          & {\CFiveTable\par{}(MIT)}                & \NATable & \DOneTable  & \DOneTable                         & \COneTable                   & \DTwoTable                       \\\hline
Mistral-7B                                & \COneTable                 & \COneTable                 & \NATable & \DOneTable                     & \DOneTable                 & \NATable         & \COneTable                         & \NATable & \NATable & \NATable                        & {\CFiveTable\par{}(Apache 2.0)}   & {\DFiveTable\par{}(Apache 2.0)}       \\\hline
Grok-1                                    & \COneTable                 & \COneTable                 & \UnkTable   & \DOneTable                     & \DOneTable                 & \UnkTable           & \COneTable                         & \NATable & \NATable & \NATable                        & \COneTable                   & \DTwoTable                       \\\hline
Phi-2                                     & \COneTable                 & \NATable                & \NATable & \DOneTable                     & \NATable                & \NATable         & \COneTable                         & \NATable & \NATable & \NATable                        & {\CFiveTable\par{}(MIT)}          & {\DFiveTable\par{}(MIT)}              \\\hline
Gemini                                    & \COneTable                 & \COneTable                 & \COneTable  & \DOneTable                     & \DOneTable                 & \DOneTable          & \COneTable                         & \COneTable  & \DOneTable  & \DOneTable                         & \COneTable                   & \DTwoTable                       

    \\\bottomrulenospacing
    
    \caption{\textbf{Model Pipeline Classification}: openness classification of components of the training, evaluation and deployment pipelines of currently available large language models. “N/A” in this table corresponds to "Not Applicable”, whereas “?” means the information is not publicly available. If a model has more than one source of code or data source for a given component, the final classification is taken by considering the strictest license. For conciseness, in the table header we use "FT" as a stand in for "Fine-Tuning".}
    \label{tab:classification_full}
    \end{longtable}
}

\clearpage
\twocolumn

\subsection{Open-source GenAI Governance}
\label{app:ai_reg}

The urgency of assaying the risks and opportunities of open-source GenAI is further underscored by recent regulatory developments around the world. The EU AI Act~\citep{AIAct} has since matured into the world's first comprehensive and enforcable regulatory framework on AI governance, and is set to introduce specific obligations to providers and deployers (users) of open-source general purpose AI models, and systems built thereon. President Biden's Executive Order on AI~\citep{house_fact_2023} is thought to significantly affect open-source developers also, and, of course, China's approach to AI regulation continuous to be governed by state intervention~\citep{china_cyber_2023, translate_interim_2023}. While these regulations may carve in stone certain aspects of future open-source GenAI governance, fundamental questions surrounding concepts such as \textit{general-purpose models of systemic risk} (EU AI Act) or \textit{dual-use foundation models} (Biden's EO) remain up to debate. Importantly, particularly in the case of the EU AI Act, many regulations have been designed to be adaptable in line with future technological progress. Our debate therefore remains highly relevant to open-source GenAI governance.

Recent years have seen the emergence of regulatory frameworks across the world that are already, or will soon, interact with the real-world governance of open-source Gen AI models. These efforts have been accompanied by increasing efforts at streamlining on the international stage, starting from 2023 G7 Hiroshima Summit and the Bletchley declaration~\citep{bletchley_2023}, and culminating in various national and transnational initiatives forming a network of AI safety institutes in the United Kingdom (UK), United States of America (US), European Union (EU), and elsewhere. Prior to the launch of ChatGPT on November 29th, 2023, such regulations were mostly targeted at (i) containing the spread of \textit{deepfakes} in order to safeguard election integrity -- e.g., the EU's 2022 amendments to the Digital Services Act --, or (ii) to exercise wider information control against the spread of ``rumors", such as the Chinese government's 2019 \textit{Regulations on the Administration of Online Audio and Video Information Services}~\citep{sheehan_chinas_2023}.
At the same time, the economic benefits of open-source AI models and systems have been almost unanimously recognized across the world. The launch of ChatGPT, and its rapid adoption among users worldwide, led policymakers to focus on general-purpose AI (GPAI) regulation.

\subsubsection{The EU AI Act}
The first \textit{comprehensive} regulatory framework governing general-purpose AI -- including provisions for open-source Gen AI -- may be the EU AI Act, which is expected to come into full force by 2026~\citep{AIAct}.
The legislation will apply to anyone putting AI services, or their outputs, on the EU market for professional purposes, while exempting recreational or academic use, as well as matters relevant to national security. It guards providers of open-source general-purpose models against risks emanating from downstream use by limiting the providers' responsibilities to a number of transparency obligations. These transparency obligations include the high-level documentation of training data provenance, as well a specification of intended use cases. 
Entities deploying Gen AI \textit{deepfakes} are required to disclose their AI-generated nature. These requirements will apply to small business owners to a lesser degree. While comprehensive, the EU AI Act will not apply to recreational or research use and will be superseded by the EU member states' individual national security interests. Open-source Gen AI providers may face additional procedures and obligations if their models are classified as \textit{general-purpose AI (GPAI) models of systemic risk}, an intentionally vaguely defined criterion that will be adapted as technology progresses. Importantly, the EU AI Act, as perhaps the EU's first transnational legislation, explicitly affirms the economic benefits of open-source AI.

\subsubsection{Biden's Executive Order}
President Biden's 2023 \textit{Executive Order (EO) on Safe, Secure, and Trustworthy Artificial Intelligence}~\citep{house_fact_2023} continues to follow a \textit{``soft law''} approach of earlier EOs, largely trading enforceable regulation for voluntary industry commitments~\citep{pricewaterhousecoopers_overview_2024}. Safety and security measures surrounding AI technology include requirements for developers to share red-teaming results with the US federal government, and for companies working on ``dual-use'' foundation models (\textit{i.e.}, systems with civilian and military applications) and/or with large compute clusters to provide regular activity reports. The National Institute of Standards and Technology (NIST) is set up to play a key role in developing standards for secure and safe AI. Instead of placing hard restrictions on the use of certain AI technology (as the EU AI Act explicitly does), Biden's EO focuses on promoting best practices, evaluations, and standard development across a wide variety of aspects including security and risk mitigation. For example, it includes references to biological weapons, AI-generated content watermarking, and labor market impacts, and, additionally, measures for attracting foreign national AI talent through streamlining visa procedures and by providing assistance to small businesses and developers. 
National security interests are also formulated, including the reporting of foreign users of US Infrastructure as a Service (IaaS) products, as well as promoting the development of AI-driven tools to detect cyber vulnerabilities.

\subsubsection{China's Gen AI Legislation} 
The earliest legal framework specifically targeting Gen AI models and systems, the Chinese government's \textit{Provisional Administrative Measures of Generative Artificial Intelligence Services (Generative AI Measures)} \citep{china_cyber_2023,translate_interim_2023}, came into force in China in August 2023. These regulations pose strict obligations on providers of Gen AI, ranging from outcome-driven provisions (e.g., requiring generative AI services to not produce illegal or untruthful content) to provenance obligations on training data and model weights, and measures targeted to protect intellectual property and privacy rights \citep{chong_china_2023}. From the point of view of open-source model developers, the inability to predict future downstream use of models and systems provided introduces legal risks that require regulatory containment. Although open-source Gen AI plays a significant role in the Chinese economy, however, these regulations do not seem to target open-source (GP)AI models specifically \citep{china_emerging_2024}. 

\subsubsection{The Middle East}
\textbf{Saudi Arabia.} 
In August 2019, as part of Saudi Arabia's Vision 2030 introduced by Crown Prince Mohammed Bin Salman, the Saudi Data and AI Authority (SDAIA) was established by a royal decree. SDAIA aims to advance this vision, with the National Center for AI serving as a key component.
Saudi Arabia, through SDAIA, has adapted and released its first version of AI ethics in September 2023 \citep{SDAIA_AI_ethics}. The document outlines Saudi's stance on AI risks, categorized from minimal to unacceptable risks with a comprehensive risk management plan covering data, algorithms, compliance, operations, legality, and regulatory risks. The AI ethics strongly supports the transparent development and deployment of AI
, reflecting that \textit{``transparent and explainable algorithms ensure that stakeholders affected by AI systems [...] are fully informed when an outcome is processed by the AI''}. Moreover, SDAIA has quickly embraced the generative AI wave. In collaboration with NVIDIA, SDAIA developed ``Allam'' \citep{Allam}, Saudi Arabia's first national-level LLM model, an Arabic LLM designed to provide summaries and answer questions, drawing information from cross-checked online sources. While Allam was closed source 
and only a beta version interface is accessible, there are still several pieces of evidences that Saudi Arabia is in favor of open-source. For instance, the Digital Government Authority \citep{digital_gov_auth} issued free and open-source government software licenses to 6 government agencies in 2022. This entails reviewing and publishing the source code ``in a way that opens the field of cooperation and unified standards among government agencies". The general directions with the laid down compliance regulations, stated principles, and open source government suggest that Saudi Arabia is in favor of open source.

\textbf{United Arab Emirates (UAE).} In October 2017, the UAE Government launched the \textit{``UAE Artificial Intelligence Strategy''} \citep{UAE_strategy_Gov}, spanning sectors from education to space. Shortly after, His Excellency, Omar Al Olama became the world's first AI minister. The UAE has been in favor of open-source in their policies, for instance, as stated in the strategy \textit{``Objective 7: Provide the data and the supporting infrastructure essential to become a test bed for AI''} and that \textit{``The UAE has an opportunity to become a leader in available open data for training and developing AI systems''}. Moreover, the strategy states that \textit{``The UAE’s ambition is to create a data-sharing program, providing shared open and standardized AI-ready data, collected through a consistent data standard''}. More recently, the UAE through the Technology Innovation Institute (TII) has open-sourced its LLM Falcon \citep{falconllm}, including its 180B parameter version, for both research and commercial use \citep{falconllm_opensource}. This all indicates the UAE’s positive take towards open-source models.

\subsubsection{AI Regulation Efforts in Other Countries}

In 2019, the Organization for Economic Co-operation and Development (OECD) introduced their AI Principles, a recommendation by the council on general-purpose AI. These principles were ratified by the G20 council, and have been adopted by at least 42 of the organization's participating countries \citep{noauthor_oecds_nodate, australian_2024}. 

Some countries have on-going legislation efforts or have issued policies specifically on Gen AI, addressing mainly sector-based issues. These include Australia \citep{noauthor_interim_nodate}, Canada \citep{noauthor_guidelines_nodate}, New Zealand \citep{kaldestad_new_2023}, Norway \citep{council2023ghost}, Singapore \citep{noauthor_mas_nodate, noauthor_first_nodate}, among others. India has published working papers on the issue of AI and enacted the Digital Personal Data Protection Act in 2023 tackling privacy issues related to Gen AI \citep{india_regulation} -- it is yet to regulate on general-purpose Gen AI and the open sourcing of models. Brazil is working on two main legislative proposals to regulate AI, one inspired in the US framework (Bill no. 21, from 2021) and another inspired on the EU framework (Bill No. 2338, from 2023), yet these do not have provisions for open-source Gen AI models. A few other countries are in the process of running public consultations on how to regulate generative AI, such as the case of Chile \citep{httpsmagnetcl_articulo_nodate} and Uruguay \citep{noauthor_mesa_nodate}. 

\end{document}